\documentclass[12pt]{spieman}  
\usepackage{amsmath,amsfonts,amssymb}
\usepackage{graphicx}
\usepackage{setspace}
\usepackage{tocloft}
\usepackage{multirow}
\title{Segmentation of Levator Hiatus Using Multi-Scale Local Region Active contours and Boundary Shape Similarity Constraint}

\author[a]{Xinling Zhang}
\author[b]{Xu Li}

\author[a]{Ying Chen}
\author[c]{Yixin Gan}
\author[b]{Dexing Kong}
\author[a]{Rongqin Zheng}
\affil[a]{Department of Medical Ultrasonic, the Third Affiliated Hospital of Sun Yat-Sen University, Guangzhou, 510000, China.}
\affil[b]{School of Mathematical Sciences, Zhejiang University, Hangzhou, 310025, China.}
\affil[c]{Department of Medical Ultrasonic, Henan Provincial People's Hospital, Zhengzhou, 450003, China. }

\cftpagenumbersoff{figure}

\cftpagenumbersoff{table}
\begin{document}
\maketitle

\begin{abstract}
  In this paper, a multi-scale framework with local region based active contour and boundary shape similarity constraint is proposed for the segmentation of levator hiatus in ultrasound images. In this paper, we proposed a multi-scale active contour framework to segment levator hiatus ultrasound images by combining the local
  region information and boundary shape similarity constraint. In order to get more precisely initializations and reduce the computational cost, Gaussian pyramid
  method is used to decompose the image into coarse-to-fine scales. A localized region active contour model is firstly performed on the coarsest scale image to get
  a rough contour of the levator hiatus, then the segmentation result on the coarse scale is interpolated into the finer scale image as the initialization.
  The boundary shape similarity between different scales is incorporate into the local region based active contour model so that the result from coarse scale
  can guide the contour evolution at finer scale. By incorporating the multi-scale and boundary shape similarity, the proposed method
  can precisely locate the levator hiatus boundaries despite various ultrasound image artifacts. With a data set of 90 levator hiatus ultrasound images,
  the efficiency and accuracy of the proposed method are validated by quantitative and qualitative evaluations (TP, FP, Js) and comparison with other two state-of-art active contour segmentation methods (C-V model, DRLSE model). The proposed model performed best among three methods with TP of $94\%\pm 2.1\%$; FP of $1.49\%\pm1.3\%$; Js of $93\%\pm1.75\%$ respectively.
\end{abstract}


{\noindent \footnotesize\textbf{*}Rongqin Zheng\footnote{Xinling Zhang and Xu Li are co-first authors. They contribute the same to the manuscript. Dexing Kong and Rongqin Zheng are corresponding authors.},  \linkable{zhengrq@mail.sysu.edu.cn}. }
\begin{spacing}{2}   

\section{Introduction}
\label{sect:intro} Pelvic floor dysfunction undermines the quality of life for a large number of women \cite{Me2000}. As one of the most widely used imaging modality to assess the anatomical integrity and    function of pelvic floor, ultrasound is simple, safe, radioactive-free and affordable to use.  Milena et al. confirm that 3D ultrasound is a reliable measurement for levator hiatus\cite{24}. Studies \cite{DietzHP2005}\cite{AbdoolZ2009}\cite{DietzHP2010}\cite{Van2014} show that the dimensions of levator hiatus have correlation with severity of prolapse, levator muscle avulsion, prolapse recurrence after surgery and muscle trauma. No matter in clinical practice or off-line analysis, recognizing the edge of levator hiatus exactly is the crucial foundation for diagnosis. Until now, the segmentation of the levator hiatus is generally performed by manually delineating. Studies indicate that inexperienced operators may fail to detect the contour in early stage\cite{25}. It means that segmentation of the levator hiatus is experience-dependent for novice, while, tedious and time- consuming.  Computer-aided automatic or semi-automatic segmentation algorithms are required to precisely detect levator hiatus boundary and can significantly decrease the burden of the technicians and the examination time.

In the past few decades, various approaches have been proposed for image segmentation. The active contour model (ACM) is one of these famous methods. It can be classified into two categories: edge- based methods and region-based methods. Edge-based active contour models \cite{M. Kass1991}\cite{C. Xu1998} use information like edges and gradients as external force to drive the contour to the desired boundaries. Unlike edge-based models, region-based models often utilize the region statistical information to guide the evolution of a contour. Comparing to edge-based methods, smoothed and weak boundaries can be handled better. Chan and Vese \cite{T.Chan2001} proposed the famous C-V model for two-phase segmentation based on the mean intensities inside/outside the contour.
Let $I:$ $ \Omega \rightarrow   R$ be the input image. C is a closed contour represented by a level set function $\phi (x) , x \in \Omega $, that is $C:= \{\ x \in \Omega | \phi(x)=0\}\ $. The region inside the contour is represented as
 $ \Omega_{in}=\{ \ x \in \Omega | \phi(x) > 0  \} $ and the region outside the contour is $\Omega_{out}= \{ \ x \in \Omega | \phi(x) < 0  \} $. Then the energy functional of the C-V model can be reformulated by the level set function: \\
     \begin{equation}
     \begin{split}
     E^{cv}(\phi,u_1,u_2)&= \mu \int_{\Omega}|\nabla H(\phi(x))|dx+\lambda_1\int_{\Omega}(I-u_1)^2 H(\phi(x)) dx\\
     &+\lambda_2\int_{\Omega}(I-u_2)^2 (1-H(\phi(x)))dx,\\
     \end{split}
     \end{equation}
where $\mu,\lambda_1 ,\lambda_2$ are fixed positive constants. The first term is the regularization term, imposing a smoothness constraint on the geometry of the contour. $u_1,u_2$ are two constants that represent the average intensities inside and outside the contour. $H(\phi(x))$ is the Heaviside function\cite{OS}.
%
The C-V model is based on the assumption that image intensity is homogeneous. However, it may fail to segment the ultrasound images. Ultrasound images usually suffer from noise, low contrast, intensity inhomogeneity, blurred or missing boundaries and other image artifacts. A lot of approaches have been proposed for ultrasound image segmentation. Employing probability density functions to model the intensity distribution of ultrasound data have been extensively studied \cite{Boukerroui2003}\cite{Sarti2005}\cite{Tao2006}. Liu et al. \cite{BLiu2010} proposed an active contour model based on the probability difference between shifted Rayleigh distribution and the intensity distribution of the ultrasound images. Huang et al. \cite{JHuang2011} proposed an active contour model based on maximum likelihood estimation with Fisher-Tippett distribution for the segmentation of ultrasound images. Besides, shape information is also greatly used in the segmentation procedure. Leventon et al. \cite{Leventon20001}
 were the first to incorporate the prior shape knowledge within the variational framework for image segmentation. They proposed a segmentation method including two steps: initial segmentation and its correction based on a shape prior model. Mahadavi et al. \cite{Mahadavi2010} introduced a parameterized tapered ellipsoid as shape constraint for prostate segmentation from TRUS. Gong et al.\cite{L.Gong2004} developed a Bayesian segmentation algorithm based on deformable super-ellipses for prostate ultrasound images. To reduce the computational costs and the risk of stuck in local minima,  multi-scale approach has been an efficient method \cite{PN}. Lin et al. \cite{L.Saroul2008} proposed a multi-scale segmentation framework based on active contour models for the segmentation of ultrasound images. However, a large number of segmentation methods have been widely explored for prostate \cite{XuLi}, thyroid \cite{Iakovidis2007}, breast\cite{Chen} etc., only a limited effort has been made in pelvic tissues.

 In this paper, we proposed a multi-scale active contour framework to segment levator hiatus ultrasound images by combining the local region information and boundary shape similarity constraint. In order to get more precisely initializations and reduce the computational cost, Gaussian pyramid method is used to decompose the image into coarse-to-fine scales. A localized region active contour model is firstly performed on the coarsest scale image to get a rough contour of the levator hiatus, then the segmentation result on the coarse scale is interpolated into the finer scale image as the initialization. The boundary shape similarity between different scales is incorporate into the local region based active contour model so that the result from coarse scale can guide the contour evolution at finer scale. By incorporating the multi-scale and boundary shape similarity, the proposed method can precisely locate the levator hiatus boundaries despite various ultrasound image artifacts. Experiments results demonstrate the advantages of the proposed model.

 This paper is organized as follows. Details processing of the proposed multi-scale segmentation framework was introduced in section 3. In section 4, we test the proposed model on real ultrasound images. Then we discussed the results and made conclusion in the last section.
\section{The proposed model}


\subsection{ Multi-scale based local region active contour with boundary shape similarity}
 There are several challenges for the segmentation of levator hiatus from ultrasound images. We show some of them in Fig.1. First, some of the levator hiatus boundaries are blurry, weak or even missing (Fig.1 A). Second, the intensity is extremely inhomogeneous inside the levator hiatus. For example, the urethra may have similar intensity performance with the muscles around the levator hiatus(Fig.1 B). Third, pixels in the background region may have similar intensity performance with the region inside the levator hiatus (Fig.1 C). In order to overcome these difficulties while segmentation, we proposed a multi-scale framework based on local image information and boundary shape similarity.
 \begin{figure*}
\begin{center}
  \includegraphics[width=2in,height=2in]{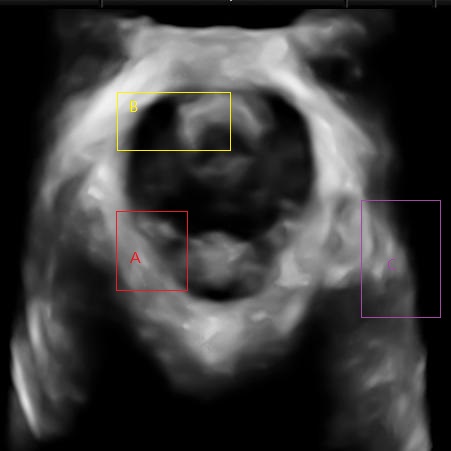}
   \caption{\footnotesize{Challenges for the segmentation of levator hiatus.}}
  \end{center}
   \end{figure*}

 \begin{figure*}
\begin{center}
  \includegraphics[width=4.5in,height=3in]{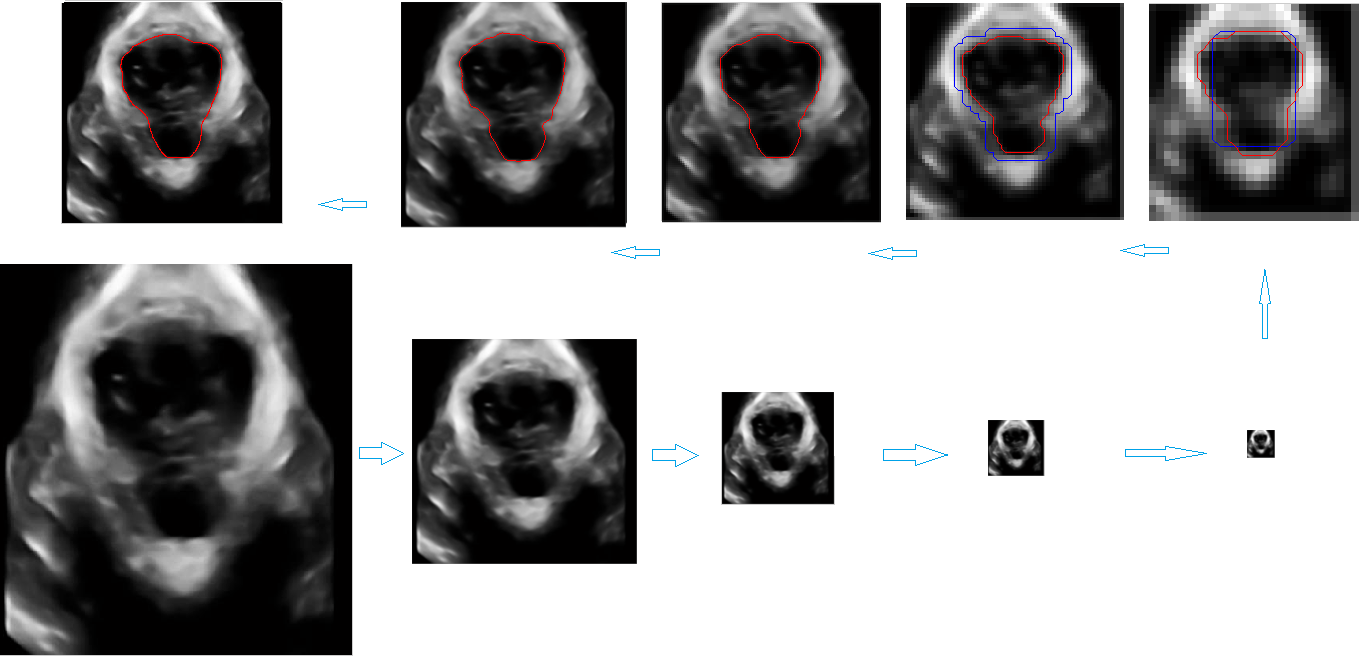}
   \caption{\footnotesize{The pipeline of the proposed model. The first image in the second row is the original image, the following images in the second row are multi-scale images generated from Gaussian pyramid. Images in the first row are segmentation results of the multi-scale images
   respectively. The blue contour is the initialization on the finer scale image which is the interpolation of the segmentation result of the coarser scale.}}
  \end{center}
   \end{figure*}
The proposed segmentation framework is based on active contour models. The overview of the proposed method is shown in Fig.2.  There are two main steps in the proposed method. In the first step, the original image is decomposed into N sub-images of different scales. We take $N=5$ in the proposed algorithm. The proposed algorithm is started at the coarsest scale. By transferring the major computation into the coarse scales, the proposed algorithm can avoid trapping into local solutions. In the second step, a local region active contour with boundary shape similarity is utilized to segment the multi-scale image serious. The segmentation result of the coarse scale image is interpolated and passed to next finer scale as initial contour and the boundary shape similarity constraint is also based on the segmentation from coarse scale. By incorporating the local region active contour and boundary shape similarity within the multi-scale framework, the proposed model can successfully segment the levator hiatus from the ultrasound images which are usually with high speckles, serious intensity inhomogeneous, weak/blurry boundaries etc.
 The total energy functional of the proposed model is: 
\begin{equation}
E=\omega_{1} E_{local}+ \omega_{2} E_{s}+ \omega_{3} R,
\end{equation}
$\omega_{1}$, $\omega_{2}$, $\omega_{3}$ are weight parameters. $E_{local}$ is a local region based energy, $E_{2}$ represents the boundary shape similarity, R is the regularization term that keeps the smoothness of the evolving contours. In this paper, $R$ is defined by:\\
\begin{center}
$R(\phi)=\int_{\Omega}|\nabla H(\phi(x))|$,
\end{center}
which is the same as in (1).
\subsection{local region energy}
Due to the intensity inhomogeneous in the ultrasound images, algorithms based on global information usually get over-segmentation results. However, if we focus on the region around the levator hiatus boundaries, over-segmentation affect can be depressed. To reduce the influence of the image intensity inhomogeneous, $E_{local}$ is defined to use the local intensity information around the evolving contours. For each point $x$, a point energy is defined as:
\begin{equation}
\begin{split}
E_{x}=\int_{y\in P_{x}}(I-u(x))^2H(\phi(x))dy+\int_{y\in P_{x}}(I-v(x))^2(1-H(\phi(x)))dy\\
\end{split}
\end{equation}
where $P_{x}$ denotes a neighborhood of x. $u(x)$ and $v(x)$ are the average intensities of the pixels in the patches inside and outside the contour respectively.
Then, a banded region around zero level set $\phi$ is defined by $B_{x}$, which has the following form:
\begin{equation}
   B_{\epsilon} (\phi(x))=\left\{
\begin{array}{lcl}
1       &      & x\in P_{x},\\
0  &      & otherwise.\\
\end{array} \right.
\end{equation}
\begin{figure*}[htbp]
\begin{centering}
   \includegraphics[width=3in,height=2in]{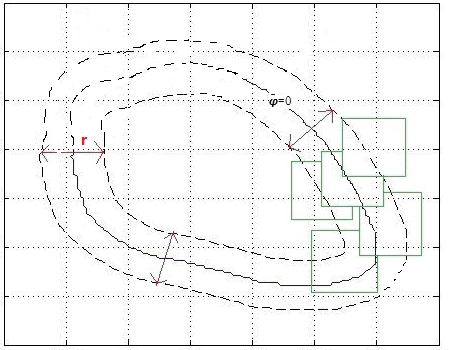}
   \caption{\footnotesize{Scheme of local energy terms. The local statistic of intensity is computed on the pixels in the region between the two dot-dashed lines.}}
   \end{centering}
   \end{figure*}
All the computation of image statistics is based on the banded region defined by $B{\epsilon}$.
Thus the local region based energy becomes:
\begin{equation}
\begin{split}
&E_{local}=\int_{\Omega}B_{\epsilon}(\phi(x))E_{x}dx\\
&=\int_{\Omega}B_{\epsilon}(\phi(x))(\int_{y\in P_{x}}(I-u(x))^2H(\phi(y))dy+\int_{y\in P_{x}}(I-v(x))^2(1-H(\phi(y)))dy)dx\\
\end{split}
\end{equation}

\subsection{Boundary shape similarity}

As explained in the previous section, the image is decomposed into N sub-images. The levator hiatus boundaries at fine scales are irregular and contain a lot of details, while the ones at coarse scales are more regular and continuous. However, these boundaries are the contours with different sizes but share commonly similar shape. To avoid boundary leakages and get more precious segmentation results at finer scales, we incorporate the boundary shape similarity between coarse and fine scales into the multi-scale framework. The proposed shape similarity does not allow large deformation of the contour during the evolution. After obtaining a rough contour from the coarse scale, the boundary shape similarity acts as an additional constraint to refine the contours at finer scales.
  In this paper, the boundary shape similarity constraint is defined as:
  \begin{equation}
  S(\phi)=\int_{\Omega}H(\phi(x)))B_s(x)dx,
  \end{equation}
where
 \begin{equation}
   B_{s} (x)=\left\{
\begin{array}{lcl}
0      &      & min_{\bar x}D(x,\bar x)<r_{s},\\
D(x,\bar x)  &      & otherwise.\\
\end{array} \right.
\end{equation}
\begin{figure*}[htbp]
\begin{centering}
   \includegraphics[width=4.5in,height=3in]{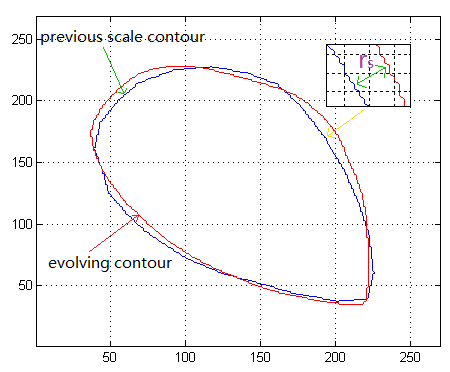}
   \caption{\footnotesize{Scheme of boundary shape similarity. The red contour is the evolving contour and the blue contour is the segmentation from previous scale.}}
   \end{centering}
   \end{figure*}
 The proposed boundary shape similarity imposed to penalize the distance between the current evolving contour and the contour from coarser scale, discourage the evolving contour from leaking a region with a weak/blurry edge or even without an edge. When the distances between the evolving contour points and the rough contour points are smaller than $r_s$, the proposed shape similarity constraint is zero. It is noted that this energy term is not used in the energy functional for the segmentation of the coarsest scale image.

Then the total energy functional of the proposed is:

\begin{equation}
\begin{split}
E&=\omega_{1} E_{local}+ \omega_{2} S+ \omega_{3} R \\
&=\int_{\Omega}B_{\epsilon}(\phi(x))(\int_{y\in P_{x}}(I-u(x))^2H(\phi(y))dy\\
&+\int_{y\in P_{x}}(I-v(x))^2(1-H(\phi(y)))dy)dx\\
&+\int_{\Omega}H(\phi(x)))B_s(x)dx+\int_{\Omega}|\nabla H(\phi(x))|\\\label{8}
\end{split}
\end{equation}
The energy functional Eq.\ref{8} can be minimized by solving the following gradient flow:
\begin{equation}
\begin{split}
&\frac{\partial \phi}{\partial t}=\omega_1(\int_{y\in P_x}B_{\epsilon}(\phi(x))\delta_{\epsilon}(\phi(y))(-(I-u(x)^2\\
&+(I-v(x))^2))dy)+\omega_2\delta_{\epsilon}(\phi(x))B_s(x)+\omega_3\delta_{\epsilon}(\phi) div(\frac{\nabla\phi}{|\nabla\phi|})\\\label{9}
\end{split}
\end{equation}
Eq.\ref{9} is embedded into the multi-scale processing to segment levator hiatus ultrasound images. At the coarsest scale, the boundary shape similarity constraint has not been used since there is no previous contour. In this case, the proposed model becomes the active contour model in a strip region around the evolving contour. However, the intensity inhomogeneous and speckle noises are reduced to a certain extent during the Gaussian pyramid decomposition, we can easily locate the rough contour of the levator hiatuses. Then at finer scales, the rough contour supervises the evolution as shape constraint. In the multi-scale framework, the levator hiatus boundary is captured continuously from coarse to fine scales. The proposed model has advantages in the following two aspects. Firstly, the images to be segmented are firstly decomposed into several scales by the Gaussian pyramid algorithm. By doing this, the intensity inhomogeneous and speckles in the coarsest scales can be relieved to some extent. Also, the total computation can be reduced. Secondly, by introducing the boundary shape similarity constraint, we can reduce the possibility of getting into incorrect locations while the contour evolving at finer scales.

\section{Experiments and Results}
The proposed algorithm is implemented with Matlab R2011a on a PC of CPU 2.5GHz, RAM 6.00G. In this section, we apply the proposed method to real levator hiatus ultrasound images. The performance of the proposed model is tested on a data set of 90 levator hiatus ultrasound images from No.3 Affiliated Hospital of Sun Yat-Sen University. The computation time is about 6s for levator hiatus ultrasound images($523\times418$). To demonstrate the advantages of the proposed model, two well-known region based active contour models: the C-V \cite{T.Chan2001} model and the distance regularized level set (DRLSE) \cite{CL}, are utilized for comparison. During our experiment,
for the C-V model, we set $\epsilon=0.01$, $\triangle t = 0.1$ and the parameter for the arc length term is: $\nu = 650.25$. For the DRSLE model, $\triangle t = 1$, $\mu=0.2$,
$\lambda=5$ and $\alpha=-3.0$. For the proposed algorithm, we set $\omega_1=0.9$, $\omega_2=0.1$ and $\omega_3=0.8$.
For quantitative analysis, we compute the following metrics:
\begin{equation}
  TP=\frac{|\Omega_{m}\cap\Omega{a}|}{|\Omega_{m}|}.
  \end{equation}
  \begin{equation}
  FP=\frac{|\Omega_{m}\cup\Omega_{a}-\Omega_{m}|}{|\Omega_{m}|}.
  \end{equation}
  \begin{equation}
  Js=\frac{|\Omega_{m}\cap\Omega{a}|}{|\Omega_{m}\cup \Omega_{a}|}.
  \end{equation}
Here $\Omega_{a}$ is the segmentation result of the proposed algorithm and $\Omega_{m}$ is the ground truth delineated by the radiologists with 5 years of experience in 2/3/4D pelvic floor ultrasound.

\begin{figure*}
\begin{center}
   \includegraphics[width=4.5in,height=3in]{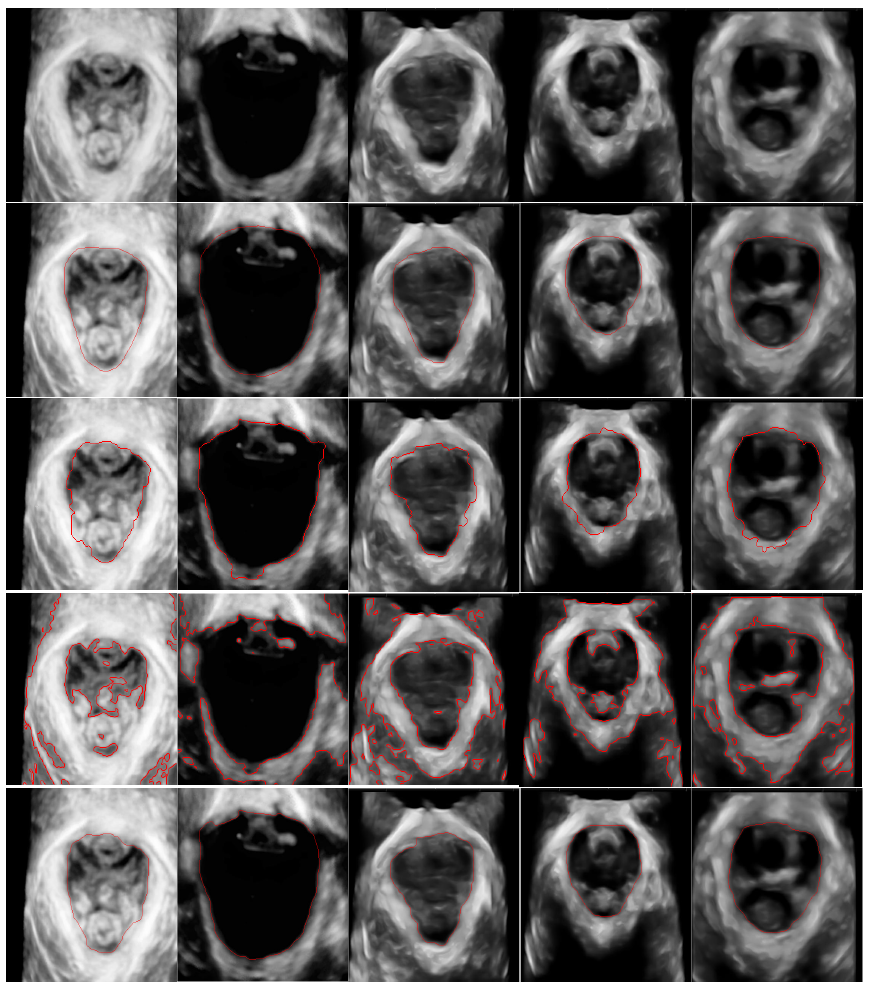}
   \caption{\footnotesize{Segmentation results of the proposed model, C-V model and the DRSLE model. (First row) The original levator hiatus ultrasound images. (Second row): manually delineated contours. (Last 3 rows) Segmentation results of the DRLSE model, C-V model and the proposed model respectively. }}
  \end{center}
   \end{figure*}
   \begin{figure*}
\begin{centering}
   \includegraphics[width=4.5in,height=3in]{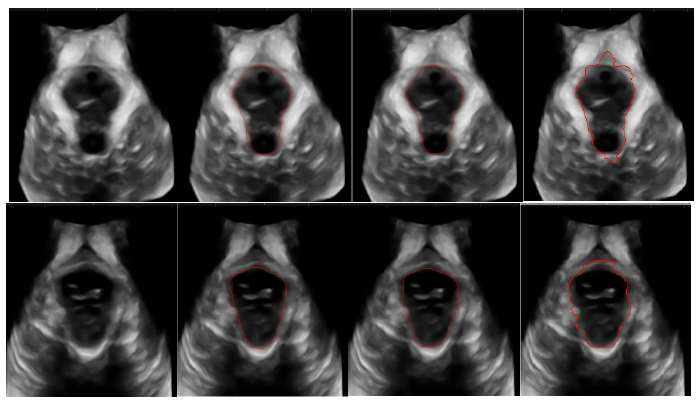}
   \caption{\footnotesize{(First column) The original images. (Second column) The ground truths. (Third column) Segmentation results of the proposed model with boundary shape similarity. (Fourth column) Segmentation results without boundary shape similarity term.}}
   \end{centering}
   \end{figure*}


\begin{figure*}
\begin{center}
\includegraphics[width=4.5in,height=3in]{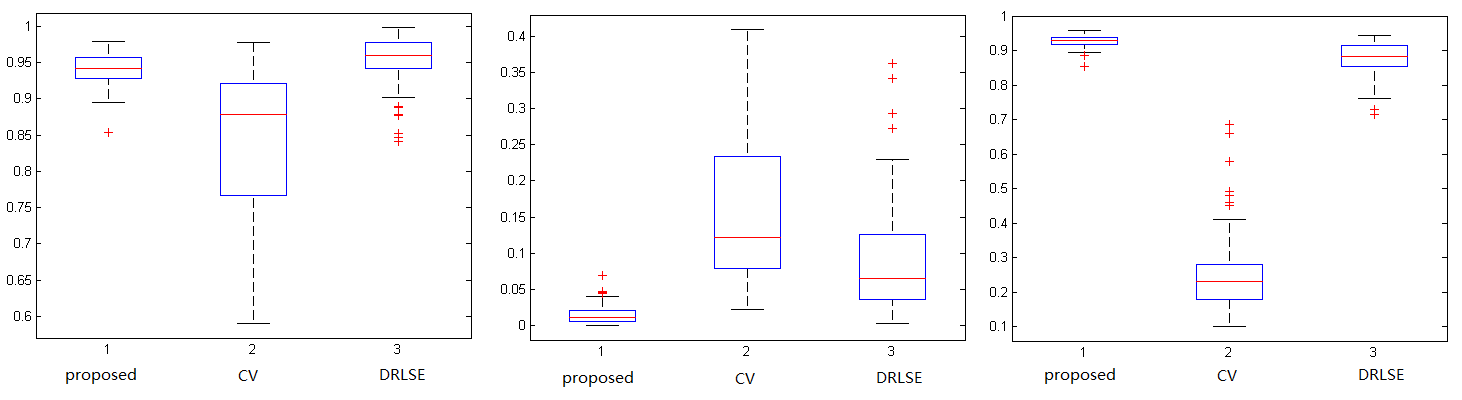}
\caption{\footnotesize{TP, FP and Jaccard similarity of the proposed model, C-V model and the DRLSE model. }}
\end{center}
\end{figure*}

In Fig.5, we show the segmentation results of the proposed model and two other active contour models(C-V model and DRLSE model). The original images and contours delineated by the radiologists are presented in the first and second rows. The segmentation results of the DRLSE, C-V and proposed models are shown in the last three rows respectively. From the segmentation results, we can see that the C-V model gets the worst segmentation results due to the presence of speckles, intensity inhomogeneity, weak/blurry boundaries in the ultrasound images. The DRLSE method performs better than C-V model, it can roughly detect the object contour. However, the contour leaks out at the weak/blurry boundaries of the levator hiatus.

In addition, although we have already obtained a rough contour of levator hiatus at the coarse scale, the contour may stuck in inaccurate locations without the boundary shape similarity constraint at finer scales. As shown in the last column of Fig.6, the final contours evolve into the fake boundary around the levator hiatus boundaries. In contrast, the proposed approach with boundary shape similarity captures the objects boundaries more precisely(the third column of Fig.6).
 Fig.7 shows the quantitative metrics of these approaches. The average TP, FP and Js of the proposed method is $94\%\pm 2.1\%$, $1.49\%\pm1.3\%$ and $93\%\pm1.75\%$ respectively. It is interesting to find that the TP values of both the DRLSE and the proposed model are very high. However, high TP values or low FP values do not mean high segmentation accuracy since both over-segmentation or under-segmentation may also lead to high TP values or low FP values, respectively. If we examine the segmentation results in details, it is easy to find that the final contours of the DRLSE method do not fit the levator hiatus boundaries accurately. The TP value of C-V is lowest and the FP value is quite high which means that parts of the background regions are wrongly counted as object regions by C-V model. The proposed approach gets the best performance. Moreover, the proposed model has the highest Js value which indicates that the contour detected by the proposed approach fits the object boundaries best among the three methods.

\section{Conclusion}

 In this study, a multi-scale based active contour model with boundary shape similarity is presented for the segmentation of levator hiatus ultrasound images. Gaussian pyramid algorithm is used to constructing a multi-scale representation for each input images. The boundary shape similarity between each scale is incorporated into the multi-scale framework to constrain the evolution of the contours. By decomposing the images into different scales, it can reduce the computation costs and get more precise initialization of the proposed algorithm. With the constraint of boundary shape similarity term, the contour can avoid leakages at weak/blurry boundaries. The proposed model is able to handle levator hiatus ultrasound images with intensity inhomogeneous, speckles and low contrast. Both qualitative and quantitative analysis show that the proposed model is superior to previous active contour models.

\subsection*{Disclosures}
The authors declared that they have no conflicts of interest to this work.
\acknowledgments
This work is supported by key project of NSF of China under grant number 11531013, key research and development project
of Zhejiang Province under grant number 2015C01028 and the fundamental research funds for the central universities.



\begin{thebibliography}{1}

\bibitem{Me2000}MacLennan, H. Alastair, et al. The prevalence of pelvic floor disorders and their relationship to gender, age, parity and mode of delivery. BJOG-Int. J. Obstet. Gy. 107.12 (2000): 1460-1470
\bibitem{24}Weinstein, M. Milena, et al. The reliability of puborectalis muscle measurements with 3-dimensional ultrasound imaging. AM J. Obstet. Gynecol. 197.1 (2007): 68-e1.
\bibitem{25}F.Siafarikas et al., Learning process for performing and analyzing 3D/4D transperineal ultrasound imaging and interobserver reliability study. Ultrasound Obst. Gyn., 41(3), (2013), p. 312-7.
\bibitem{DietzHP2005}H.P.Dietz, C. Shek, B. Clarke. Biometry of the pubovisceral muscle and levator hiatus by threedimensional pelvic floor ultrasound. Ultrasound Obst. Gyn., 25, (2005), 580-5.
\bibitem{AbdoolZ2009}Z. Abdool, KL. Shek, HP. Dietz. The effect of levator avulsion on hiatal dimension and function. AM J. Obstet. Gynecol. , 201, (2009), 89.e1-e5.
\bibitem{DietzHP2010}H.P.Dietz, V. Chantarasorn, KL. Shek . Levator avulsion is a risk factor for cystocele recurrence. Ultrasound Obst. Gyn., 36, (2010), 76-80.
\bibitem{Van2014}GA. Van Veelen, KJ. Schweitzer, K. van Delft K, KB. Kluivers, M. Weemhoff, CH. van der Vaart. Diagnosing levator avulsions after first delivery by tomographic ultrasound: reliability between observers from different centers. Int Urogynecol J, 25, (2014), 1501-1506.
\bibitem{M. Kass1991}M. Kass, A. Witkin, D. Terzopoulos, Snake: active contours model, Int. J. Comput. Vis. 1(4), (1991), 1167-1186.
\bibitem{C. Xu1998}C. Xu and J. L. Prince, Snakes, Shapes, and Gradient Vector Flow, IEEE Trans. Image Process., 7(3), (1998), 359-369.
\bibitem{T.Chan2001}T. Chan, L. Vese, Active contours without edges, IEEE Trans. Image Process. 10 (2), (2001), 266-277.
\bibitem{Boukerroui2003}D. Boukerroui, A. Baskurt, JA. Noble, O. Basset. Segmentation of ultrasound images-multiresolution 2D and 3D algorithm based on global and local statistics. Pattern Recognition Lett. 24(4¨C5), (2003), 779-990.

\bibitem{Sarti2005}A. Sarti, C. Corsi, E. Mazzini, C. Lamberti. Maximum likelihood segmentation of
ultrasound images with Rayleigh distribution. IEEE Trans. Ultrason., 52(6), (2005), 960-974.

\bibitem{Tao2006}Z. Tao, HD. Tagare, JD. Beaty. Evaluation of four probability distribution models
for speckle in clinical cardiac ultrasound images.IEEE Trans. Med. Imaging; 25(11), (2006), 1483-1491.
\bibitem{BLiu2010}B. Liu, H. D. Cheng, J. Huang, et al. Probability density difference-based active contour for ultrasound image segmentation. Pattern Recogn., 43(6), (2010), 2028-2042.
\bibitem{JHuang2011}J. Huang, X. Yang, Y. Chen, A fast algorithm for global minimization of maximum likelihood based on ultrasound image segmentation,  Inverse Probl. Imag., vol.5, (2011), 645-657.
\bibitem{Leventon20001}M. E. Leventon, O. Faugeras, W.E.L.Grimson, et al. Level set based segmentation with intensity and curvature priors, IEEE Workshop on Mathematical Methods in Biomedical Image Analysis, (2000), 4-11.
 \bibitem{Mahadavi2010}S.S.Mahdavi, N. Chng, I.Spadinger et al. Semi-automatic Segmentation for Prostate Interventions. Med. Image Anal., 15(2), (2010), 226-37.
 \bibitem{L.Gong2004}L. Gong, S. D. Pathak, D. R. Haynor, P. S. Cho, Y. Kim. Parametric shape modeling using deformable superellipses for prostate segmentation. IEEE Trans. Med. Imaging, 23(3), (2004), 340-349.
 \bibitem{PN}N. Paragios, R.Deriche.Coupled geodesic active regions for image segmentation: a level set approach.In: lecture notes in computer science,vol.1843.Springer; (2000), 224-240.
 \bibitem{L.Saroul2008}N.Lin, W. Yu, J.S.Duncan. Combinative multi-scale level set framework for echocardio graphic image segmentation. Med. Image Anal.;7(4), (2003), 529-537.
 \bibitem{OS}S. Osher and J. A. Sethian. Fronts propagating with curvature-dependent speed: algorithms based on Hamilton-Jacobi formulations. J. Comput. Phys., 79(1), (1998), 12-49.
 \bibitem{XuLi}X. Li, C. Li, A. Fedorov, T. Kapur and X. Yang. Segmentation of prostate from ultrasound images using level sets on active band and intensity variation across edges. Med. Phy., 43(6), (2016), 3090-3103.
 \bibitem{CL}C. Li, C. Xu, C. Gui, and M. D. Fox, Distance Regularized Level Set Evolution and its Application to Image Segmentation, IEEE Trans. Image Processing, vol. 19 (12), (2010), 3243-3254.
\bibitem{Iakovidis2007}D.K. Iakovidis, M. Savelona, S.A. Karkanis, et al. A genetically optimized level set approach to segmentation of thyroid ultrasound images. Appl. Intell., 27(3), (2007), 193-203.
\bibitem{Chen}D.R. Chen, R.F. Chang, W.J. Wu, et al. 3-D breast ultrasound segmentation using active contour model. Ultrasound Med. Biol., 29(7), (2003), 1017-1026.



\end{thebibliography}


%

\listoffigures
\listoftables

\end{spacing}
\end{document}